\newcommand{\CLASSINPUTinnersidemargin}{18mm}
\newcommand{\CLASSINPUToutersidemargin}{12mm}
\newcommand{\CLASSINPUTtoptextmargin}{20mm}
\newcommand{\CLASSINPUTbottomtextmargin}{25mm}
\documentclass[10pt,conference,a4paper]{IEEEtran}

\usepackage[utf8]{inputenc}

\usepackage{times}

\usepackage{graphicx}
\usepackage{subfigure}
\DeclareGraphicsExtensions{.png,.eps,.ps,.pdf}

\usepackage{url}

\usepackage[spanish,es-tabla]{babel}

\begin{document}

\title{Aportes para el cumplimiento del Reglamento (UE) 2024/1689 en robótica y sistemas autónomos}


\author{\IEEEauthorblockN{Francisco J. Rodríguez Lera, Yoana Pita Lorenzo, David Sobrín Hidalgo, Laura Fernández Becerra}
\IEEEauthorblockA{Universidad de León\\
Grupo de Robótica, EIIIA, Campus de Vegazana, S/N, 24071, León\\
fjrodl@unileon.es, \{ypital00,inflfb00, dsobrh00\}@estudiantes.unileon.es}
 \and
 \IEEEauthorblockN{Irene González Fernández, Jose Miguel Guerrero Hernández}
 \IEEEauthorblockA{Universidad Rey Juan Carlos,\\
 Intelligent Robotics Lab, EIF, Fuenlabrada, 28943, Madrid, Spain\\
\{irene.gonzalezf,josemiguel.guerrero\}@urjc.es}
}

\maketitle

\begin{abstract}
La ciberseguridad en robótica se erige como un aspecto clave dentro del Reglamento (UE) 2024/1689, también conocido como Ley de Inteligencia Artificial, que establece directrices específicas para sistemas inteligentes y automatizados. Una distinción fundamental en este marco regulatorio es la diferencia entre robots con Inteligencia Artificial (IA) y aquellos que operan mediante sistemas de automatización sin IA, dado que los primeros están sujetos a requisitos de seguridad más estrictos debido a su capacidad de aprendizaje y autonomía. En este trabajo, se analizan herramientas de ciberseguridad aplicables a sistemas robóticos avanzados, con especial énfasis en la protección de bases de conocimiento en arquitecturas cognitivas. Además, se propone un listado de herramientas básicas para garantizar la seguridad, integridad y resiliencia de estos sistemas, y se propone un caso práctico, enfocado al análisis de la gestión del conocimiento del robot, donde se definen diez criterios de evaluación que permitan asegurar la conformidad con el reglamento y reduzcan los riesgos en entornos de interacción humano-robot (HRI).
\end{abstract}


{\bf Tipo de contribución:}  {\it Investigación original}

\section{Introducción}

La ciberseguridad en robótica es un aspecto que cobra especial importancia en el marco del Reglamento (UE) 2024/1689, conocido como \textit{The AI Act} \cite{EURegulation2024},  el cual establece directrices específicas para sistemas automatizados e inteligentes. Una distinción fundamental dentro de este contexto es la diferencia entre robots que integran Inteligencia Artificial (IA) y aquellos que operan mediante sistemas de automatización sin IA. Esta diferenciación impacta directamente en los requisitos de seguridad y cumplimiento normativo, dado que los robots con IA están sujetos a regulaciones más estrictas en función de sus capacidades y nivel de autonomía.

La normativa europea de ciberseguridad adopta un enfoque basado en la evaluación del riesgo, lo que significa que las medidas de seguridad aplicadas a cada sistema dependen de su grado de exposición a amenazas. Por ejemplo, un robot que realiza acciones consideradas críticas, conducción autónoma o entretenimiento en escuelas de infantil, se enfrenta a riesgos significativamente mayores que un robot doméstico, lo que exige niveles de protección diferenciados.

Este enfoque flexible permite equilibrar la innovación con la seguridad, evitando la imposición de requisitos innecesarios para sistemas de bajo riesgo, al tiempo que se refuerzan las protecciones en entornos críticos. En este contexto, el uso de herramientas de auditoría de ciberseguridad es clave para garantizar la seguridad de los sistemas robóticos y su cumplimiento con las normativas vigentes.

Este trabajo analiza los riesgos de ciberseguridad que afectan a los robots autónomos en el contexto del Reglamento (UE) 2024/1689, identificando las amenazas específicas que surgen de la regulación existente y su impacto en la robótica cognitiva. En particular, se presenta un enfoque sistemático para la evaluación de la base de conocimiento de un robot con arquitectura cognitiva, abarcando modelos de IA, ficheros de configuración y datos biométricos.

La metodología propuesta permite evaluar aspectos críticos de seguridad, incluyendo identificación de riesgos, protección de datos sensibles, gestión de actualizaciones y parches, resiliencia ante ciberataques, cumplimiento normativo, gestión de vulnerabilidades, planificación de respuesta a incidentes y auditoría continua. Este enfoque facilita la integración de medidas de seguridad en sistemas autónomos, garantizando su conformidad con las normativas europeas y fortaleciendo su capacidad de operar en entornos de interacción humano-robot.

A través de la integración de estos mecanismos de seguridad, se garantiza la integridad y confiabilidad de la toma de decisiones en sistemas robóticos, evitando modificaciones no autorizadas en la base de conocimiento y fortaleciendo la resiliencia de la Interacción Humano-Robot (HRI) ante amenazas cibernéticas. Este trabajo proporciona un marco de referencia para el desarrollo de sistemas cognitivos robóticos seguros y confiables.  

El artículo se compone de la siguiente estructura, donde, la sección \ref{sec:estadodelarte} revisa el estado del arte en ciberseguridad aplicada a la robótica, analizando estudios sobre la seguridad en sistemas como ROS \cite{quigley2009ros} y ROS 2 \cite{macenski2022robot} y destaca la importancia de proteger la base de conocimiento de los robots mediante medidas técnicas que permitan identificar, mitigar y gestionar vulnerabilidades, asegurando así la integridad y resiliencia de los datos en su arquitectura cognitiva. La sección \ref{sec:cumpliminetoreglamento} detalla las directrices establecidas por el Reglamento  para asegurar la seguridad y fiabilidad de los sistemas autónomos e inteligentes, clasificando los sistemas robóticos con  IA desplegada según su nivel de riesgo y estableciendo requisitos específicos. La sección \ref{sec:riesgosrobots} aborda la seguridad en sistemas robóticos con IA desplegada mediante un enfoque multinivel, similar al Robot Security Framework, donde se destacan los aspectos clave de cada capa y se integra la aplicación de consideraciones éticas y regulatorias. Por otra parte, la sección \ref{sec:herrauditorias} menciona algunas de las herramientas de ciberseguridad específicas para sistemas de automatización, control industrial y robótica, donde se detallan sus principales funciones. La sección \ref{sec:casopract} analiza los niveles funcionales de una arquitectura híbrida en robots autónomos, e identifica los desafíos de seguridad que esta integración plantea en el contexto de la IA y su cumplimiento con el Reglamento (UE) 2024/1689. La sección \ref{sec:criteval} analiza la gestión del conocimiento en un robot cognitivo basado en ROS 2, identificar riesgos de seguridad y proponer estrategias de mitigación, sentando las bases para un marco de análisis integral en seguridad de IA aplicada a robótica.
Finalmente la sección \ref{sec:conclusiones} de Conclusiones revisa todos los elementos revisados en el artículo y la propuesta como punto de partida para futuros trabajos. 

\section{Estado del Arte}
\label{sec:estadodelarte}

Estudios  como los planteados en \cite{sedgewick2014framework,cybersecurity2018framework} han establecido bases fundamentales en la seguridad en infraestructuras críticas, en particular a través de marcos de ciberseguridad como el desarrollado por el Instituto Nacional de Estándares y Tecnología (NIST). Este marco proporciona un enfoque estructurado para evaluar, mitigar y gestionar los riesgos de ciberseguridad en sistemas altamente conectados y tecnológicamente avanzados. A través de modelos como el \textit{Cybersecurity Framework}, se establecen mecanismos flexibles y adaptativos para la identificación, protección, detección, respuesta y recuperación ante ciberamenazas en infraestructuras clave, incluyendo la robótica. Nuestra investigación se apoya en estos marcos para extender su aplicación a la seguridad de la base de conocimiento del robot, considerando la especificidad de los sistemas de IA utilizados en este contexto.

Un primer enfoque relevante es el de \cite{vilches2021introducingrobotsecurityframework}, que analiza la seguridad en el contexto de Robot Operating System (ROS) y su evolución hacia ROS 2. Aunque proporciona una visión técnica sólida, su tratamiento de la inteligencia artificial en el robot es limitado. Nuestra contribución en este sentido es ampliar el modelo propuesto, integrando la seguridad de la base de conocimiento en los sistemas deliberativos y subsimbólicos del robot.

Por otro lado, estudios como \cite{braga2024hacia} abordan la necesidad de construir sistemas robóticos con IA desplegada confiables, centrándose en ciberseguridad, ética y regulación. Sin embargo, algunas de estas aproximaciones tienden a enfocarse exclusivamente en el aspecto ético, sin profundizar en los desafíos técnicos. En nuestro trabajo, abordamos directamente la dimensión técnica, proporcionando una evaluación estructurada de las vulnerabilidades y medidas de protección para la base de conocimiento del robot.

Desde una perspectiva más cercana a la toma de decisiones, \cite{Basan2019} plantea un modelo que profundiza en este aspecto, aunque no enfatiza la importancia del conocimiento previo que posee el robot. Nuestro trabajo complementa esta aproximación, asegurando que la seguridad de la base de conocimiento se integre de manera efectiva en la arquitectura cognitiva del robot.

Por otra parte, ~\cite{oruma2024architectural} presenta el SecuRoPS Framework, un modelo arquitectónico detallado para la integración de robots en entornos públicos. Aunque este marco incluye una perspectiva de seguridad amplia, nuestra contribución específica se centra en la protección del conocimiento del robot, asegurando la integridad y resiliencia de los datos almacenados y procesados tanto en la infraestructura local como en la nube.

\section{Cumplimiento del Reglamento (UE) 2024/1689 en Robótica Autónoma para Interacción Humano-Robot}
\label{sec:cumpliminetoreglamento}

El Reglamento (UE) 2024/1689~\cite{EU_AI_Regulation_2024,BOE_2024} establece directrices específicas para la seguridad y confiabilidad de los sistemas autónomos e inteligentes. En el contexto de la interacción humano-robot (HRI), los requisitos clave incluyen la clasificación del riesgo del sistema de IA, la implementación de medidas de ciberseguridad, y el cumplimiento de estándares de supervisión y transparencia.

El reglamento clasifica los sistemas robóticos con IA desplegada en diferentes niveles de riesgo, determinando los requisitos que deben cumplir:
\begin{itemize}
    \item Riesgo inaceptable: Sistemas robóticos con IA desplegada que representan una amenaza directa para la seguridad, los derechos fundamentales o el bienestar de las personas. Estos sistemas están prohibidos e incluyen:
    \begin{itemize}
        \item Manipulación y engaño perjudiciales basados en IA.
        \item Explotación de vulnerabilidades con fines perjudiciales.
        \item Puntuación social y predicción del riesgo de infracción penal.
        \item Raspado masivo de datos biométricos para bases de reconocimiento facial.
        \item Identificación biométrica remota en tiempo real para fines policiales en espacios públicos.
        \item Reconocimiento de emociones en lugares de trabajo e instituciones educativas.
        \item Categorización biométrica para inferir características protegidas.
    \end{itemize}
    
    \item Sistemas de alto riesgo: Robots con IA que pueden afectar la seguridad física o los derechos fundamentales de las personas. Ejemplos incluyen:
    \begin{itemize}
        \item Componentes de IA en infraestructuras críticas (como vehículos autónomos).
        \item IA en robots desplegados en instituciones educativas.
        \item Componentes de seguridad en productos médicos y cirugía asistida por robot.
        \item IA aplicada a la identificación biométrica y el reconocimiento de emociones.
    \end{itemize}
     \item Sistemas de riesgo limitado: Requieren transparencia en su funcionamiento, pero no presentan un riesgo crítico.
    \item Sistemas de riesgo mínimo: Pueden operar con menor supervisión, siempre que no interfieran en la seguridad o privacidad de los usuarios. Los robots de entretenimiento, sin comportamientos asociados a la personalización, podrían entrar aquí. 
\end{itemize}

Si el robot autónomo se clasifica como de alto riesgo, el cumplimiento del reglamento exige la implementación de medidas estrictas:
\begin{itemize}
    \item Gestión de riesgos: Identificación y mitigación de vulnerabilidades potenciales en el sistema robótico con IA desplegada.
    \item Documentación técnica: Mantenimiento de registros detallados sobre el desarrollo, operación y seguridad del sistema.
    \item Registro de datos: Almacenamiento de eventos relevantes durante el funcionamiento del robot para auditoría y trazabilidad.
    \item Supervisión humana: Mecanismos que permitan intervención humana en caso de fallos o riesgos de seguridad.
    \item Transparencia y explicabilidad: Información clara sobre las capacidades, limitaciones y decisiones del sistema autónomo.
    \item Robustez y seguridad: Protección ante errores, ataques adversariales y manipulación de datos.
\end{itemize}

Además de la regulación sobre la IA, el reglamento exige garantizar la seguridad física y la protección contra ciberataques:
\begin{itemize}
    \item Seguridad del hardware: Implementación de controles físicos, como cerraduras y monitoreo de acceso al robot.
    \item Protección contra software malicioso: Supervisión continua y detección de amenazas en la infraestructura del robot.
    \item Gestión de identidad y acceso: Autenticación robusta para restringir el acceso a los sistemas internos.
    \item Seguridad de redes: Uso de VPNs y cifrado para proteger la comunicación entre el robot y la nube.
    \item Protección de datos: Implementación de mecanismos de cifrado y controles de acceso para la información almacenada.
    \item Auditorías y supervisión: Implementación de sistemas SIEM y análisis periódicos de vulnerabilidades.
\end{itemize}

Si el robot utiliza servicios en la nube para procesamiento o almacenamiento de datos, el reglamento exige:
\begin{itemize}
    \item Seguridad en la nube: Implementación de cifrado y acceso restringido a datos sensibles.
    \item Monitoreo de acceso: Registro de accesos y detección de anomalías en la gestión del conocimiento.
    \item Auditoría de conformidad: Evaluación periódica para garantizar la alineación con los requisitos normativos.
\end{itemize}

El cumplimiento de estas normativas exige un monitoreo y verificación constantes de la seguridad en los sistemas autónomos. Para ello, las auditorías de ciberseguridad juegan un papel fundamental, ya que permiten detectar vulnerabilidades, evaluar la resiliencia de los sistemas ante ataques y garantizar el cumplimiento de los requisitos establecidos en el Reglamento (UE) 2024/1689.


\section{Riesgos en la IA de los Robots}
\label{sec:riesgosrobots}
La seguridad en sistemas robóticos con IA debe abordarse desde múltiples capas, siguiendo una estructura similar al Robot Security Framework\cite{vilches2021introducingrobotsecurityframework}. A continuación, se presentan los principales aspectos a considerar en cada capa y vamos a contribuir teniendo en cuenta la aplicación de consideraciones éticas y regulatorias.

\subsection{Capa Física}
Aunque la IA es predominantemente un componente de software, su integración puede afectar la seguridad física del robot. Se deben considerar:
\begin{itemize}
    \item \textbf{Protección del hardware}: Protección de hardware especializado (TPU, GPU) contra manipulaciones.
    \item \textbf{Resguardo de sensores}: Evitar sabotajes o interferencias físicas en sensores utilizados por la IA.
    \item \textbf{Redundancia en percepción}: Implementación de múltiples sensores para mejorar la resiliencia a fallos.
\end{itemize}

\subsection{Capa de Red}
El uso de IA en robots implica la transmisión de datos de entrenamiento e inferencias a través de redes, lo que introduce riesgos de seguridad:
\begin{itemize}
    \item \textbf{Cifrado de modelos y datos}: Protección de modelos y datos de entrenamiento en tránsito y en reposo.
    \item \textbf{Defensa contra ataques adversariales en la red}: Prevención de ataques de envenenamiento de datos y manipulación de inferencias.
    \item \textbf{Autenticación de dispositivos}: Implementación de autenticación fuerte para evitar accesos no autorizados.
\end{itemize}

\subsection{Capa de Firmware}
La IA en robots a menudo se ejecuta en hardware embebido con firmware especializado:
\begin{itemize}
    \item \textbf{Ejecución en entornos confiables (TEE)}: Uso de entornos de ejecución seguros para evitar manipulación de modelos.
    \item \textbf{Actualización segura de modelos}: Verificación de integridad y autenticidad en actualizaciones de firmware y modelos de IA.
\end{itemize}

\subsection{Capa de Aplicación}
La IA impacta significativamente la seguridad en la capa de aplicación, donde se gestionan los modelos, inferencias y decisiones del robot:
\begin{itemize}
    \item \textbf{Explicabilidad y Transparencia}: Implementación de técnicas como LIME o SHAP para auditar decisiones de IA.
    \item \textbf{Defensas contra ataques adversariales}: Entrenamiento robusto y validaciones de entrada para evitar manipulación de inferencias.
    \item \textbf{Monitoreo de comportamiento}: Implementación de detección de drift y auditoría de modelos en producción.
    \item \textbf{Privacidad y protección de datos}: Cumplimiento con GDPR y otros marcos regulatorios.
    \item \textbf{Control de acceso a modelos}: Restricción del acceso a modelos y API de inferencias mediante autenticación robusta.
    \item \textbf{Cifrado en comunicaciones}: Protección de inferencias transmitidas a la nube o a otros dispositivos.
\end{itemize}

\subsection{Consideraciones Éticas y Regulatorias}
Además de la seguridad técnica, la integración de IA en robots plantea cuestiones éticas y legales:
\begin{itemize}
    \item \textbf{Cumplimiento con estándares internacionales}: Asegurar conformidad con normativas como NIST AI RMF, ISO/IEC 23894.
    \item \textbf{Responsabilidad y trazabilidad}: Registro de decisiones críticas y análisis de fallos.
    \item \textbf{Implementación de mecanismos de seguridad}: Implementación de \textit{kill switch} en caso de mal funcionamiento.
\end{itemize}

\section{Herramientas de auditorías de ciberseguridad }
\label{sec:herrauditorias}
Algunas de las herramientas de auditoría de ciberseguridad específicas para sistemas de automatización, control industrial y robótica incluyen:

\begin{itemize}
    \item \textbf{Sistemas de Detección de Intrusiones (IDS) y Prevención (IPS)}: Monitorean el tráfico de red en busca de actividades sospechosas y pueden bloquear ataques en tiempo real. 
    
    \item \textbf{Firewall de Próxima Generación (NGFW)}: Proporcionan una capa adicional de seguridad al inspeccionar el tráfico de red y aplicar políticas de seguridad avanzadas.

    \item \textbf{Soluciones de Gestión de Información y Eventos de Seguridad (SIEM)}: Recopilan y analizan datos de seguridad de múltiples fuentes para detectar y responder a incidentes de seguridad.

    \item \textbf{Autenticación Fuerte y Control de Acceso}: Implementar autenticación multifactor y políticas de control de acceso estrictas para proteger los sistemas críticos. 
    \textit{Ejemplo}: RSA SecurID, Okta.
    
    \item \textbf{Segmentación de Redes}: Dividir la red en segmentos más pequeños para limitar el alcance de un posible ataque. 
    \textit{Ejemplo}: Cisco TrustSec, VMware NSX.
    
    \item \textbf{Sistemas de Gestión de Parches}: Herramientas que ayudan a mantener los sistemas actualizados con los últimos parches de seguridad. 
    \textit{Ejemplo}: Microsoft System Center Configuration Manager (SCCM), Ivanti Patch Management.
    
    \item \textbf{Monitorización y Detección de Amenazas}: Soluciones que permiten la monitorización continua de los sistemas para detectar y responder a amenazas en tiempo real. 
    \textit{Ejemplo}: Darktrace, CrowdStrike Falcon.
    
    \item \textbf{Soluciones de Respuesta y Recuperación ante Incidentes}: Herramientas que ayudan a gestionar y mitigar los efectos de un incidente de seguridad. 
    \textit{Ejemplo}: CyberArk, FireEye.
\end{itemize}

Estas herramientas son esenciales para proteger los sistemas de automatización, control industrial y robótica contra ciberataques y garantizar la seguridad y continuidad de las operaciones.


\section{Caso Práctico}
\label{sec:casopract}
\begin{figure*}[t]
	\centering{
		\includegraphics[width=\textwidth]{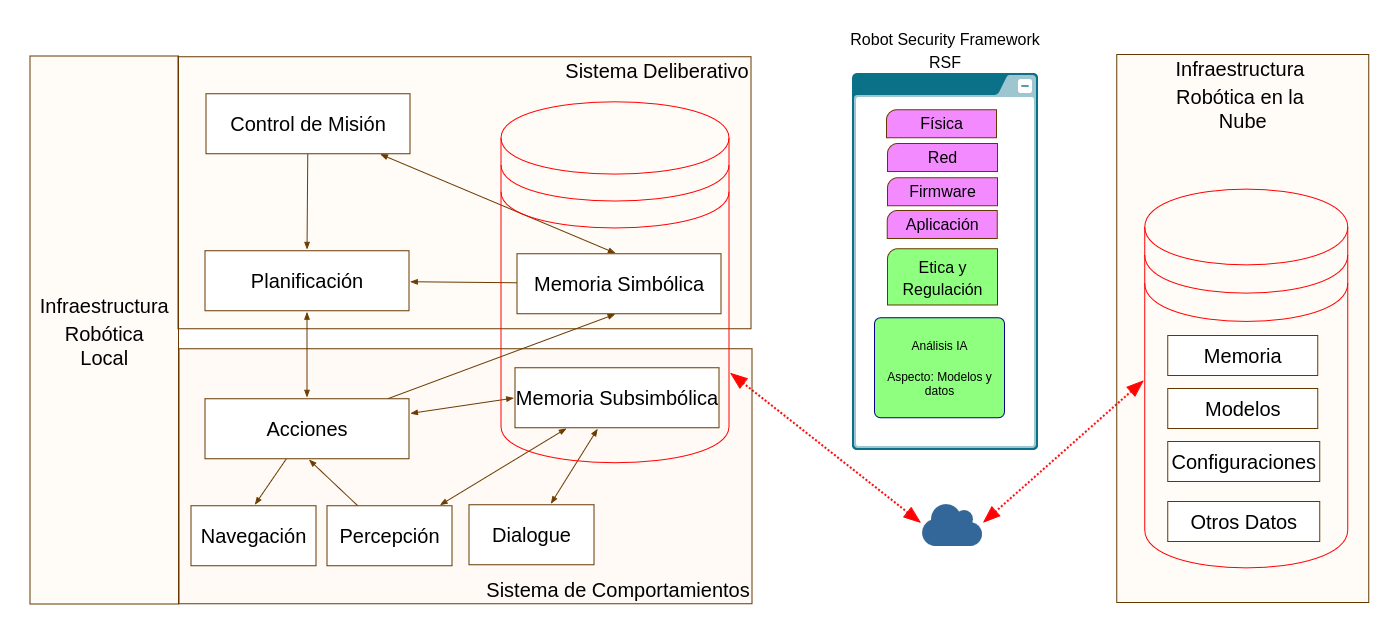}
	}
	\caption{Ejemplo de arquitectura cognitiva para robots con análisis RSF y con nuestra contribución para el análisis del conocimiento.}
	\label{fig:ArquitecturaCognitiva}
\end{figure*}

El esquema presentado en la Figura~\ref{fig:ArquitecturaCognitiva} representa una arquitectura híbrida en la que un robot autónomo opera de manera local, integrando capacidades deliberativas y de comportamiento, mientras mantiene una interacción con servicios en la nube. Esta conexión permite optimizar la toma de decisiones y la gestión del almacenamiento de datos, facilitando el acceso a modelos, configuraciones y otros recursos computacionales. Sin embargo estas aproximaciones abren nuevas problematicas en la gestión de problemas de seguridad asociados a IA. 

Las arquitecturas cognitivas en robótica están diseñadas para dotar a los robots de capacidades avanzadas de percepción, toma de decisiones y adaptación al entorno. En un esquema como el presentado, la arquitectura suele dividirse en varios niveles funcionales:

\begin{itemize}
    \item Sistema Deliberativo: Este nivel está compuesto por módulos como el control de misión y 
    la planificación, que permiten al robot generar planes de acción a largo plazo en función de los objetivos y el estado del entorno. Se apoya en una memoria simbólica, que almacena representaciones de alto nivel del conocimiento del robot, facilitando la planificación basada en modelos y el razonamiento lógico.
\end{itemize}

\begin{itemize}
    \item Sistema de Comportamientos: Ejecuta acciones en tiempo real, reactivas, basadas en la planificación enviada desde la capa deliberativa.     Incluye módulos como navegación, percepción y diálogo, que permiten la interacción con el entorno y con usuarios humanos.
        Utiliza una memoria subsimbólica, encargada de gestionar representaciones de bajo nivel derivadas de la percepción sensorial y la experiencia adquirida durante la operación.
\end{itemize}
    
\begin{itemize}
    \item     Interacción con la Nube:   La conexión con servicios en la nube permite que el robot acceda a modelos de aprendizaje, configuraciones y otros datos sin necesidad de mantener toda la información localmente.    Facilita la actualización y mejora continua de sus capacidades cognitivas mediante el aprendizaje federado o la descarga de modelos entrenados previamente en entornos similares.
        
\end{itemize}
    
La implementación de una arquitectura híbrida en robots autónomos~\cite{penmetcha2020smart,chen2021fogros}, que combine procesamiento local con servicios en la nube, plantea desafíos significativos en el contexto del Reglamento (UE) 2024/1689 sobre inteligencia artificial. A continuación, se destacan las principales problemáticas:

\begin{itemize}
    \item     Clasificación de Riesgo: El reglamento establece una clasificación de los sistemas robóticos con IA desplegada basada en el nivel de riesgo que representan para los derechos fundamentales y la seguridad. Los sistemas de alto riesgo están sujetos a requisitos más estrictos. Las arquitecturas híbridas deben evaluarse para determinar su nivel de riesgo y cumplir con las obligaciones correspondientes. 

    \item     Transparencia y Explicabilidad: El reglamento exige que los sistemas robóticos con IA desplegada sean transparentes y explicables, permitiendo a los usuarios comprender cómo se toman las decisiones. En una arquitectura híbrida, donde parte del procesamiento ocurre en la nube, garantizar esta transparencia puede ser complejo debido a la posible opacidad de los procesos en la nube.

    \item     Supervisión Humana: Es fundamental que los sistemas de IA permitan la supervisión humana efectiva. En arquitecturas híbridas, la supervisión puede verse dificultada por la distribución de componentes entre el dispositivo local y la nube, lo que podría afectar la capacidad de intervención humana en tiempo real. 

    \item     Ciberseguridad: La conexión a la nube introduce riesgos adicionales de ciberseguridad. El reglamento requiere que los sistemas robóticos con IA desplegada implementen medidas para protegerse contra ataques que puedan comprometer su funcionamiento o la seguridad de los datos. Las arquitecturas híbridas deben abordar estos riesgos de manera integral. 

    \item     Gestión de Datos y Privacidad: El reglamento enfatiza la protección de datos y la privacidad. En una arquitectura híbrida, los datos pueden transferirse entre el robot y la nube, lo que plantea desafíos en términos de control y protección de la información personal. 

    \item     Responsabilidad y Rendición de Cuentas: Determinar la responsabilidad en caso de fallos o daños es más complejo en sistemas híbridos debido a la participación de múltiples actores (desarrolladores, proveedores de servicios en la nube, operadores, etc.). El reglamento exige una clara asignación de responsabilidades. 
\end{itemize}

En este artículo, nos enfocaremos en la gestión del conocimiento del robot, específicamente en los mecanismos de información asociados a lo componentes de memoria y la infraestructura de datos en la nube, tal como se representa en la figura. Para ello, utilizaremos el enfoque presentado en este estudio\cite{vilches2021introducingrobotsecurityframework}.

\section{Criterios de Evaluación y Análisis de Amenazas}
\label{sec:criteval}
Para este análisis, consideraremos la gestión del conocimiento en un robot cognitivo basado en ROS 2, centrándonos en los mecanismos de almacenamiento y acceso a la información tanto en la memoria local como en la infraestructura en la nube. En este contexto, el robot utiliza una Base de Conocimiento (\textit{Knowledge Base}), que se asocia a las memorias presentadas en la figura anterior y que integra múltiples fuentes de conocimiento:

\begin{itemize} 

\item \textbf{Memoria simbólica}: Contiene representaciones explícitas, legibles y manipulables mediante reglas lógicas o lenguajes formales. Se emplea principalmente en la planificación deliberativa y la toma de decisiones de alto nivel. Un ejemplo típico es el uso de lenguajes como \textit{PDDL}, que permiten definir acciones, objetivos y restricciones del entorno de forma estructurada.
\item \textbf{Memoria subsimbólica}: Gestiona representaciones implícitas, codificadas mediante parámetros de modelos entrenados (por ejemplo, redes neuronales profundas). Esta forma de conocimiento es esencial para tareas perceptivas, motoras y de aprendizaje adaptativo. Incluye:
\begin{itemize}
    \item \textbf{Modelos de visión por computador}: como \texttt{YOLO\_ROS}~\cite{yoloros2023}, utilizados para la detección de objetos en tiempo real a partir de imágenes de cámara, integrando modelos de Ultralytics en sistemas robóticos.
    \item \textbf{Modelos de audio y voz}: por ejemplo, \texttt{Whisper\_ROS}~\cite{whisperros2023}, utilizado para reconocimiento de voz en entornos ruidosos o multimodales.
\end{itemize}

\end{itemize}

Estos son los sistemas de memoria clásicos, a los que tenemos que añadir también lo que se podría considerar como Memoria híbrida. Corresponde a sistemas que, aunque están implementados con estructuras subsimbólicas, operan funcionalmente sobre contenido simbólico. Este es el caso de los modelos de lenguaje de gran escala (\textit{LLMs}), como los utilizados en \texttt{Llama\_ROS}~\cite{llamaros}, que permiten generar, completar o interpretar instrucciones en lenguaje natural. Estos modelos actúan como interfaces simbólicas, pero con una base computacional subsimbólica (p.e. embeddings) lo que les confiere un carácter mixto que requiere una atención especial en cuanto a seguridad, explicabilidad y trazabilidad del conocimiento generado.

Estas memorias,  y sus  modelos asociados, presentan múltiples vulnerabilidades que pueden comprometer la seguridad del sistema. Existen ataques dirigidos tanto a modelos simbólicos como subsimbólicos, afectando la toma de decisiones y la fiabilidad del conocimiento del robot. Según aproximaciones como las propuestas por OWASP\cite{john2025owasp,owasp_llm_top10}.
Los principales riesgos incluyen la manipulación de datos de entrada para alterar el comportamiento del modelo, el envenenamiento de datos en el entrenamiento y la inferencia de información sensible a partir de modelos entrenados. Además, la alteración de salidas y la explotación de vulnerabilidades en la cadena de suministro representan amenazas significativas que pueden comprometer la seguridad del conocimiento almacenado.

En el contexto de modelos de lenguaje y aprendizaje profundo, los ataques incluyen la inyección de instrucciones, la manipulación de respuestas generadas y la fuga de información sensible. Estos riesgos se ven agravados por la dependencia de modelos preentrenados y la falta de sanitización en la gestión de datos. Asimismo, el robo de modelos y el uso de técnicas adversariales pueden comprometer la confidencialidad e integridad del conocimiento del robot, tanto en su infraestructura local como en la nube.

Estos riesgos afectan directamente la integridad y confiabilidad de la base de conocimiento del robot, tanto en su infraestructura local como en la nube. Algunos efectos críticos incluyen:
\begin{itemize}
    \item Corrupción del conocimiento simbólico y subsimbólico: La manipulación de reglas PDDL, dominio o problema) o la corrupción de datos de configuración pueden generar decisiones erróneas.
    \item Compromiso de modelos de Machine Learning: Los ataques adversariales pueden alterar el rendimiento de modelos de visión como YOLO, afectando la percepción del entorno.
    \item Vulnerabilidad de LLMs y embeddings: Los modelos de lenguaje pueden ser manipulados para generar respuestas maliciosas o revelar información sensible.
    \item Riesgos en la infraestructura en la nube: El acceso no autorizado o la alteración de modelos almacenados en la nube pueden comprometer la seguridad global del sistema.
\end{itemize}

Para mitigar estos riesgos, es necesario adoptar estrategias de protección y evaluación en cada vertiente del conocimiento del robot. Algunas de las medidas clave incluyen:
\begin{itemize}
    \item Cifrado y autenticación en la base de conocimiento; uso de LUKS, GPG para ficheros y alternativas para proteger configuraciones y modelos en disco.
    \item Firmas digitales y control de versiones: para garantizar la integridad de modelos de Machine Learning y LLMs.
    \item Auditorías y supervisión continua: implementación de SIEM como Wazuh para monitorear eventos de seguridad en la infraestructura local y en la nube.
    \item Técnicas de robustez en Machine Learning uso de aprendizaje adversarial para mejorar la resistencia de modelos como YOLO ante ataques.
    \item Monitoreo del acceso en la nube: evaluación de accesos a modelos y configuraciones para prevenir manipulaciones.
\end{itemize}

Esta evaluación del estado del arte en seguridad de IA aplicada a robótica establece la base para un marco de análisis integral, como se presenta en los criterios de evaluación desarrollados en este trabajo.

Esta evaluación del estado del arte en seguridad de IA aplicada a robótica establece la base para un marco de análisis integral, como se presenta en los criterios de evaluación desarrollados en este trabajo. Recordamos que en este caso, nos hemos enfocado en la gestión del conocimiento.

\subsection{Criterio 1: Memoria Simbólica}

\textbf{Objetivo}: Evaluar la estructura y coherencia de representaciones simbólicas en planificación y razonamiento lógico.

\textbf{Justificación}: 
Los modelos simbólicos como PDDL o las ontologías permiten la planificación basada en reglas explícitas y estructuradas. Sin embargo, pueden presentar problemas de expresividad, escalabilidad y robustez ante incertidumbre en entornos dinámicos. Tanto el dominio como el problema, ficheros básicos de funcionamiento de PDDL, puede ser modificado en tiempo de ejecución y el problema

\textbf{Método}: 
\begin{enumerate}
    \item Analizar la ontología utilizada (clases, relaciones, reglas).
    \item Evaluar la consistencia lógica del modelo mediante pruebas de inferencia.
    \item Medir la eficiencia del planificador en diferentes escenarios.
    \item Examinar la integración del conocimiento simbólico con representaciones subsimbólicas.
    \item Identificar riesgos de manipulación de reglas en PDDL que puedan afectar la toma de decisiones.
\end{enumerate}

\subsection{Criterio 2: Memoria Subsimbólica}

\textbf{Objetivo}: 
Analizar la fiabilidad y precisión de los datos de configuración y parámetros almacenados. 

\textbf{Justificación}: 
Los robots dependen de archivos de configuración para parametrizar su comportamiento. Datos mal calibrados o corruptos pueden comprometer su rendimiento y seguridad. Por ejemplo, ficheros de configuración de Nav2

\textbf{Método}: 
\begin{enumerate}
    \item Verificar la coherencia y formato de los ficheros de configuración.
    \item Evaluar la sensibilidad del sistema a errores en los parámetros.
    \item Medir el impacto de variaciones en los datos de entrada sobre el comportamiento del robot.
    \item Implementar estrategias de detección de corrupción de datos.
    \item Aplicar técnicas de cifrado y control de acceso en bases de conocimiento mediante herramientas como \textbf{SROS2} para asegurar que solo los nodos autorizados accedan a esos datos, controlando el acceso al servicio o nodo que los maneja; cifrados clásicos sobre discos usando LUKS; de BBDD utilizando SQLlite; cifrados GPG sobre ficheros; uso de  SealFs~\cite{guardiola2023sealfsv2} para garantizar que los datos no han sido manipulados o sistemas basados en BLockChain para salavaguardar el estado a lo largo del tiempo en sistemas en la nube.
\end{enumerate}

\subsection{Criterio 3: Sistemas de Aprendizaje}

\textbf{Objetivo}: 
Examinar la robustez, precisión y posibles vulnerabilidades en modelos entrenados para percepción.

\textbf{Justificación}: 
Los modelos de Machine Learning son esenciales para la percepción del entorno en robots. Sin embargo, son susceptibles a ataques adversariales, sesgo en los datos y problemas de interpretabilidad.

\textbf{Método}: 
\begin{enumerate}
    \item Evaluar la precisión del modelo en distintos escenarios.
    \item Identificar sesgos en los datos de entrenamiento.
    \item Analizar la resiliencia ante ataques adversariales.
    \item Medir el impacto del modelo en la toma de decisiones del robot.
    \item Implementar firmas digitales para garantizar la integridad de los modelos, por ejemplo los de YOLO.
    \item Aplicar técnicas de aprendizaje adversarial para fortalecer modelos de visión.
\end{enumerate}

\subsection{Criterio 4: Generative AI}

\textbf{Objetivo}: 
Evaluar la capacidad de generalización, interpretación y susceptibilidad a ataques adversariales en modelos basados en representaciones densas y aprendizaje profundo. De este modo se cubre las nuevas opciones de Embeddings y Large Language Models.

\textbf{Justificación}: 
Los modelos de lenguaje y embeddings mejoran la capacidad del robot para interpretar información no estructurada. Sin embargo, pueden ser manipulados mediante inyección de instrucciones, presentar sesgos y carecer de interpretabilidad.

\textbf{Método}: 
\begin{enumerate}
    \item Evaluar la coherencia y precisión de las respuestas generadas.
    \item Identificar vulnerabilidades ante ataques de inyección de instrucciones.
    \item Medir la capacidad de adaptación a diferentes dominios.
    \item Analizar el impacto del modelo en la autonomía del robot.
    \item Aplicar firmas digitales para garantizar la integridad de modelos de LLMs.
    \item Implementar mecanismos de fallback para generar respuestas seguras en LLMs.
\end{enumerate}

\subsection{Criterio 5: Seguridad y Robustez en el Almacenamiento y Recuperación del Conocimiento}

\textbf{Objetivo}: 
Evaluar la protección del conocimiento ante corrupción, accesos no autorizados y pérdidas de información.

\textbf{Justificación}: 
El conocimiento almacenado en robots puede ser un objetivo de ataques o verse comprometido por fallos técnicos. La seguridad y robustez del almacenamiento son esenciales para garantizar la fiabilidad operativa.

\textbf{Método}: 
\begin{enumerate}
    \item Analizar los mecanismos de control de acceso y autenticación.
    \item Evaluar estrategias de recuperación ante pérdida de datos.
    \item Medir la resiliencia frente a corrupción de información y ataques externos.
    \item Aplicar medidas de sanitización en datos de entrenamiento de LLMs.
    \item Implementar auditorías para garantizar el cumplimiento de normativas como GDPR y NIST.
\end{enumerate}

\subsection{Criterio 6: Evaluación en Entornos Reales}

\textbf{Objetivo}: 
Validar el desempeño del conocimiento en escenarios operativos.

\textbf{Justificación}: 
Un sistema cognitivo puede comportarse bien en simulaciones pero fallar en entornos reales debido a variaciones no previstas. Evaluar el desempeño en el mundo real es crucial.

\textbf{Método}: 
\begin{enumerate}
    \item Comparar el rendimiento del sistema en simulaciones y pruebas reales.
    \item Evaluar la adaptabilidad ante condiciones adversas.
    \item Medir el impacto de la calidad del conocimiento en la eficacia operativa del robot.
    \item Implementar detección de ataques adversariales en modelos de visión.
    \item Realizar pruebas de penetración y simulaciones de ataques en entornos de desarrollo.
    \item Establecer mecanismos de rollback para restaurar versiones seguras de los modelos.
\end{enumerate}

\subsection{Criterio 7: Auditorías y Supervisión en Infraestructura Local}

\textbf{Objetivo}: 
Garantizar el monitoreo continuo y la validación de eventos de seguridad en la infraestructura local del robot.

\textbf{Justificación}: 
Es necesario supervisar el conocimiento almacenado y los procesos que acceden a él para detectar anomalías y vulnerabilidades que puedan comprometer la seguridad operativa del robot.

\textbf{Método}: 
\begin{enumerate}
    \item Integrar SIEM (Wazuh) para monitoreo continuo de eventos en la Knowledge Base local.
    \item Definir métricas de seguridad para evaluar la estabilidad del sistema en el robot.
    \item Monitorear las actualizaciones manuales o automáticas asociadas a los modelos y datos: desarrolladores, modelos de fuentes públicas (HuggingFace). 
    \item Implementar auditorías periódicas en los sistemas de almacenamiento y procesamiento de conocimiento.
\end{enumerate}

\subsection{Criterio 8: Auditorías y Supervisión en Infraestructura en la Nube}

\textbf{Objetivo}: 
Supervisar el acceso, integridad y seguridad del conocimiento almacenado en la infraestructura en la nube.

\textbf{Justificación}: 
Los sistemas en la nube presentan riesgos adicionales debido a la exposición a ataques externos, la dependencia de servicios de terceros y la necesidad de garantizar la integridad del conocimiento.

\textbf{Método}: 
\begin{enumerate}
    \item Monitorear el acceso y la integridad de modelos y datos almacenados en la nube.
    \item Implementar auditorías periódicas para validar la integridad del conocimiento en entornos cloud.
    \item Monitorear las actualizaciones manuales o automáticas asociadas a los modelos y datos: desarrolladores, modelos de fuentes públicas (HuggingFace). 
    \item Evaluar el cumplimiento de normativas de ciberseguridad en la gestión del conocimiento en la nube.
\end{enumerate}

\subsection{Criterio 9: Gestión de Vulnerabilidades}

\textbf{Objetivo}: 
Identificar, mitigar y gestionar vulnerabilidades en el conocimiento y los modelos utilizados por el robot.

\textbf{Justificación}: 
Las vulnerabilidades en la base de conocimiento pueden ser explotadas para manipular el comportamiento del robot, comprometer la seguridad del sistema y afectar la toma de decisiones.

\textbf{Método}: 
\begin{enumerate}
    \item Integrar herramientas de escaneo de vulnerabilidades en ROS 2 y su Sistema Operativo.
    \item Implementar listas blancas y control de acceso a los sistemas de gestión del  conocimiento del robot (cualquiera de las memorias).
    \item Validar parches de seguridad antes de su implementación.
    \item Establecer planes de respuesta ante vulnerabilidades detectadas.
\end{enumerate}

\subsection{Criterio 10: Protección de Datos Sensibles y Biometría}
\textbf{Objetivo:} Garantizar la seguridad, privacidad y cumplimiento normativo en el almacenamiento, procesamiento y acceso a datos biométricos y otra información sensible dentro de la base de conocimiento del robot.

\textbf{Justificación} El almacenamiento de datos biométricos dentro del robot y en la nube introduce riesgos de privacidad, posible uso indebido y requerimientos de cumplimiento normativo, especialmente con el Reglamento (UE) 2024/1689 y normativas como el GDPR\cite{GDPR_2016}. La exposición de estos datos puede generar vulnerabilidades críticas, por lo que su gestión debe garantizar cifrado, acceso restringido y auditoría continua.

\textbf{Método:}
\begin{itemize}
    \item Cifrado de datos en reposo y en tránsito: Aplicar técnicas de cifrado de extremo a extremo en los datos biométricos almacenados en la base de conocimiento y transmitidos a la nube (TLS 1.3).
    \item Gestión de identidad y control de acceso: Implementar autenticación multifactor (MFA) y roles de acceso estrictos para evitar accesos no autorizados a los datos biométricos.
    \item Monitorización y auditoría: Integrar sistemas de detección de accesos no autorizados (SIEM) y auditorías periódicas para detectar posibles filtraciones de datos biométricos.
    \item Minimización y anonimización de datos: Aplicar principios de minimización de datos, evitando el almacenamiento innecesario de información biométrica y, cuando sea posible, anonimizando los datos.
    \item Cumplimiento normativo: Verificar que los procedimientos de gestión de datos biométricos cumplan con el Reglamento (UE) 2024/1689, el GDPR y otras regulaciones de privacidad aplicables.
    \item Seguridad en la nube: Asegurar que cualquier dato biométrico recibido   y almacenado en la nube siga estándares de protección como cifrado avanzado (AES256) y  acceso restringido (a IAM (Identity and Access Management)).
\end{itemize}

\section{Conclusiones}
\label{sec:conclusiones}

Este estudio establece un marco de referencia integral para la ciberseguridad en sistemas robóticos con IA desplegada, alineado con el Reglamento (UE) 2024/1689. En este artículo se han abordado aspectos críticos como la identificación de riesgos, la protección de datos, la gestión de actualizaciones, la resiliencia ante ciberataques y el cumplimiento de estándares de seguridad. Históricamente, herramientas como Secure Robot Operating System 2 (SROS2), han sido clave para garantizar comunicaciones cifradas, control de acceso robusto y verificación de integridad mediante firmas digitales. Sin embargo, en el nuevo entorno es necesario integrar estrategias que se adapten a las nuevas áreas de enfoque de la Inteligencia Artificial como la IA generativa, aprendizaje automático, etc, que permitan detectar ataques adversariales en visión artificial, monitorizar amenazas por anomalías y aplicar buenas prácticas en los robots que las despliegan.

Estas medidas refuerzan la integridad y confiabilidad de los sistemas robóticos, evitando accesos no autorizados y fortaleciendo su resiliencia frente a amenazas cibernéticas. Además, promueven un equilibrio entre la innovación tecnológica y el cumplimiento normativo, consolidando un enfoque responsable para el desarrollo de sistemas seguros y adaptables. 

En este contexto, se propone un conjunto de criterios de evaluación para la validación empírica que permita  confirmar la efectividad de las herramientas, en gran parte heredado de otros entornos de ciberseguridad, y queda abierto al análisis pormenorizado de las estrategias a adoptar, siendo necesario un estudio más profundo para cada uno de los criterios generales aquí propuestos.

\section*{Agradecimientos}

Proyecto DMARCE - PID2021-126592OB-C21, PID2021-126592OB-C22  financiado por MICIU/AEI/10.13039/501100011033 y por FEDER, UE. Plan de Recuperacion, Transformación y Resiliencia, financiado por 
la Unión Europea (Next Generation) gracias al proyecto TESCAC (Trazabilidad y Explicabilidad en Sistemas Autonomos para la mejora de la Ciberseguridad) concedido por INCIBE a la Universidad de León. David Sobrín Hidalgo agradece a la Universidad de León la beca proporcionada para financiar sus estudios de doctorado. Proyecto CORESENSE, con financiación del Programa de Investigación e Innovación Horizonte Europa de la Unión Europea (Acuerdo de Subvención n.º 101070254).

\bibliographystyle{IEEEtran}
 \bibliography{jnic2025}

\end{document}